\documentclass[letterpaper, 10 pt, conference]{ieeeconf}  

\IEEEoverridecommandlockouts                              
\usepackage{amsmath,amssymb,amsfonts}
\usepackage{algpseudocode}
\usepackage{graphicx}
\usepackage{textcomp}
\usepackage{graphicx}
\usepackage{textcomp}
\usepackage{xcolor}
\usepackage{verbatim}
\usepackage{lipsum}
\usepackage{cleveref}
\Crefname{figure}{Fig.}{Figs.}
\usepackage{algorithm}

\usepackage[
    natbib=true,
    style=numeric,
    sorting=ynt
]{biblatex}
\usepackage{float}
\usepackage{tabularx}
\usepackage{booktabs}
\usepackage{array}
\usepackage{caption,subcaption}
\usepackage{siunitx}
\usepackage{float}
\usepackage{makecell}
\usepackage{amsmath}
\usepackage{amssymb}
\def\BibTeX{{\rm B\kern-.05em{\sc i\kern-.025em b}\kern-.08em
    T\kern-.1667em\lower.7ex\hbox{E}\kern-.125emX}}

\title{\LARGE \bf
On Designing Consistent Covariance Recovery from a Deep Learning Visual Odometry Engine
}

\author{Jagatpreet Singh Nir$^{1}$, Dennis Giaya$^{1}$ and Hanumant Singh$^{1}$
\thanks{$^{1}$All authors are from NU Field Robotics lab, Department of Electrical and Computer Engineering, Northeastern University, Boston, USA %
        {\tt\small \{nir.j, giaya.d and ha.singh\}@northeastern.edu}}%
}

\begin{document}

    \maketitle
\thispagestyle{empty}
\pagestyle{empty}

\begin{abstract}

Deep learning techniques have significantly advanced in providing accurate visual odometry solutions by leveraging large datasets. However, generating uncertainty estimates for these methods remains a challenge. Traditional sensor fusion approaches in a Bayesian framework are well-established, but deep learning techniques with millions of parameters lack efficient methods for uncertainty estimation.

This paper addresses the issue of uncertainty estimation for pre-trained deep-learning models in monocular visual odometry. We propose formulating a factor graph on an implicit layer of the deep learning network to recover relative covariance estimates, which allows us to determine the covariance of the visual odometry (VO) solution. We showcase the consistency of the deep learning engine's covariance approximation with an empirical analysis of the covariance model on the EUROC datasets to demonstrate the correctness of our formulation.

\end{abstract}

\section{INTRODUCTION}
The last decade highlights an increasing trend to use mobile robotic platforms beyond controlled settings \cite{skydio, waymo}. When interacting with unstructured environments in the real world, a robot must operate effectively with an incomplete and uncertain worldview. In such scenarios, mistakes can lead to potentially catastrophic results, compromising the mission's safety and endangering human lives, e.g., in driverless cars.  Consequently, researchers and designers of robotic systems have become increasingly focused on their safety and reliability\cite{kendall2017uncertainties}. One step towards ensuring safety and reliability is to quantify the uncertainty of their perception, planning, and control algorithms.

Research in Simultaneous Localization and Mapping (SLAM), a core navigation technology of an intelligent mobile robot, has been revolutionized by the application of Deep Neural Network (DNN) based SLAM frameworks \cite{bloesch2018codeslam, Czarnowski_deep_factors,teed2021droid,zihan_niceslam, zhou_deeptam, teed_deepv2d}. Such frameworks have improved dense map accuracy \cite{bloesch2018codeslam} and the ability to track in challenging environments \cite{teed_deepv2d, teed2021droid, tartanvo2020corl}. Traditional SLAM frameworks have systematic methods to quantify the estimation uncertainty of their outputs\cite{ORBSLAM3_TRO, vins_mono, Kimera} in an online sense. Sensor errors are computed using a model-based approach grounded in the geometrical and physical properties of the sensors\cite{poddar2017comprehensive, song2013survey, wang2021survey} and then minimized in a MAP estimation problem \cite{isam, g2o}.  \cite{sunderhauf2018limits}? 

\begin{figure}[h!]
    \centering
    \captionsetup{font=footnotesize, labelfont={bf, sf}}
    \includegraphics[width=\columnwidth]{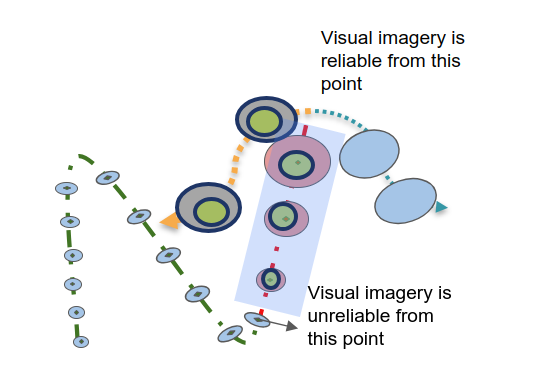}

    \caption{Covariance recovery from SLAM has utilities in many upstream tasks such as loop closure, data association, map merging, and active sensing }
    \label{fig:enter-label}
\end{figure}

However, current deep learning SLAM techniques lack such uncertainty quantification methods; uncertainty quantification in learning-based SLAM is an open problem. The deployment of learning-based SLAM requires the answer to a critical question: How trustworthy are deep-learning SLAM predictions? Traditional approaches of computing the first-order augmented Jacobians of sensor measurements are not directly applicable to learning-based SLAM \cite{kaess2009covariance}. These approaches quickly become infeasible as the network size becomes huge, typically in the order $10^9$ parameters. The predictive uncertainty originates from the data and the parameters of the neural network. For example, deep learning does not allow for uncertainty representation in regression settings, and classification models often give normalized score vectors, which do not
necessarily capture model uncertainty \cite{kendall2017uncertainties}. Accurate quantification of multivariate uncertainty will allow for the full potential of deep learning SLAM \cite{ ummenhofer2017demon, bloesch2018codeslam, teed_deepv2d, teed2021droid} to be integrated more safely and reliably in field-robotic applications.

Modern SLAM systems use various sensors, like  Inertial Measurement Units (IMUs), stereo cameras, and Light Detection and Ranging (LiDAR) systems, for robust motion tracking. Estimating the predictive uncertainty of deep learning frameworks can improve the upstream task for integrating their output with a heterogeneous sensor suite in a probabilistic framework \cite{prob_rob, g2o, isam}.

This work presents a novel covariance recovery design methodology for a pre-trained end-to-end Visual Odometry (VO) pipeline by formulating a factor graph to express the network's implicit layers. We present an approach to quantify uncertainty in terms of recovering marginal covariances of the system state from deep-learning VO engines that have implicit layers as part of their architecture. Finally, we evaluated our covariance recovery methodology on the public EUROC\cite{euroc} dataset and showed the trends of marginal covariance are consistent with the observed data. The \footnote{\url{https://github.com/jpsnir/droid_slam_covariance_models}}{code} of our covariance model is publicly available

\section{Previous work}
Uncertainty quantification for DNN regression and classification networks has been dealt with systematically in these works \cite{russell2021multivariate, hullermeier2021aleatoric, zhan-2021-uncer-quant}. Kendall \cite{kendall2017uncertainties} elucidates in his work that the predictive uncertainty of a DNN can be modeled in terms of two sources: aleatoric uncertainty (or data uncertainty) and epistemic uncertainty (or model uncertainty). It highlights that aleatoric (or data) uncertainty stems from the inherent variability in the data and its collection process and is, therefore, irreducible. In contrast, epistemic uncertainty is due to a gap in the knowledge of the model generating the data. Epistemic uncertainty typically increases when a deep learning model is presented with Out-of-distribution test samples that are not trained on.

In Bayesian deep learning models, aleatoric uncertainty estimates are obtained by either placing distributions over model weights or by learning a direct mapping to probabilistic outputs \cite{hullermeier2021aleatoric}. On the other hand, epistemic uncertainty is much harder to model. In a seminal work by Gal \cite{gal2016dropout}, Monte-Carlo dropout sampling was introduced to estimate model uncertainty by placing a Bernoulli distribution over the network's weights using a dropout layer during inference and computing expectations for mean and variance.

Approaches in deep learning-based SLAM have also applied similar principles of determining aleatoric and epistemic uncertainty \cite{gal2016uncertainty}\cite{gal2016dropout} \cite{kendall2017uncertainties}. Loquercio presented a general framework for training a deep end-to-end model \cite{loquercio2020general} for obtaining uncertainty, applied to computer vision and control tasks. Katherine \cite{katherince_covest} presents a new method to predict covariance of the sensor data given the raw sensor data when it is corrupted with dynamic actors in the scene\cite{yang2020d3vo}. Another class of neural-network design is based on the philosophy of \emph{learning to optimize} networks \cite{teed_deepv2d, teed2021droid}. These architectures have outperformed previous designs of end-to-end deep learning frameworks in SLAM. The main constituent of the \emph{learning to optimize} networks is their update operator, along with an implicit layer. The \emph{implicit layers} \cite{teed_thesis} define a differentiable layer in terms of satisfying some joint conditions as in \eqref{eq:con1}  and \eqref{eq:con2}.
\begin{align}
    y &\ni  g(x, y) = 0 \label{eq:con1}\\ 
    x &\ni \operatorname*{argmin}_x y = \sum_i^{N} e_i(x) \label{eq:con2}
\end{align} 
on its input $x$ and output $y$. Complex operations such as solving optimizing problems within a neural network become possible with such a mechanism of constraints in the network layers. Many works in SLAM have incorporated implicit layers \cite{tang_ba_net, gradslam, amos2017optnet} to improve the performance of predictions; however, none of them have focused on obtaining uncertainty measures using these constraints. 

\begin{table}[h!]
\setlength\tabcolsep{4pt} 
    \captionsetup{font=footnotesize, labelfont={bf,sf}}
    \scriptsize
    \begin{tabular}{c|c|c|c|c}
        \toprule
        S.No. & DNN framework & Application & Uncertainty & Implicit layers  \\
        \midrule
        1. & Demon\cite{ummenhofer2017demon} & dense depth   &  no  & no  \\ 
        
        2. & DroidSLAM \cite{teed2021droid} & pose + dense depth  &  no   & yes   \\
         
        3. & D3VO \cite{yang2020d3vo}  & pose + dense depth  &  yes (depth) & no    \\
        4. & Sigma-fusion \cite{sigma_fusion} & dense depth & yes (depth) & yes      \\

        6. &  NICE SLAM\cite{zihan_niceslam} & pose + dense depth & no & no   \\

        7. &  Code SLAM\cite{bloesch2018codeslam} & pose + dense depth &  yes (depth) & no    \\

        8. &  Tartan VO\cite{tartanvo2020corl} & pose & no & no    \\

        9. &  NeRF SLAM\cite{nerf_slam} & pose + depth & yes (depth) & yes    \\
        10 & DeepV2D \cite{teed_deepv2d} & depth & no & yes\\
        \bottomrule
    \end{tabular}
    \caption{Deep learning SLAM frameworks and their comparison. Note that Sigma-fusion and NeRF SLAM use DROID SLAM as their VO engine, over which they developed their improved depth mapping pipeline.}
    \label{tab:dnn_framework}
\end{table}
In \cref{tab:dnn_framework}, a summary of state-of-the-art deep-learning SLAM frameworks qualitatively presents the different aspects of their design, and their approach to quantifying uncertainty is discussed. Our work builds on using the concept of \emph{implicit layers} to recover uncertainty estimates regarding the local covariance of the regressed pose from the network. Our work more closely aligns with the ideas and formulation of \cite{sigma_fusion}, but we are estimating the covariances of poses on the manifold space to be used for navigation and sensor fusion instead of the depth uncertainty to improve the volumetric depth map. 


 This paper aligns with the theme of estimating uncertainty from a deep learning SLAM framework. Our approach to recovering marginal covariances of DNN SLAM's pose predictions depends on the use of implicit layers in a trained network\cite{teed2021droid} as opposed to learning the aleatoric uncertainty or determining epistemic uncertainty from the statistics of monte-carlo dropout sampling \cite{gal2016dropout} from an end-to-end deep neural network. 
\section{Methodology and Formulation}
\subsection{Covariance of a deep neural network in SLAM using implicit layers as constraints}
The key intuition behind our work is that when neural networks have constraints from implicit layers as depicted in \eqref{eq:con1} and \eqref{eq:con2}, the error in prediction during training, the network learns its weights to both satisfy these constraints and minimize the loss function error. The prediction uncertainty can be measured during inference as a function of the deviation from these constraints. During network training, the implicit layers affect the update of network weights through backpropagation such that the network learns to understand the constraints from an implicit layer. 
However, during inference, the same implicit layers impose constraints that need to be obeyed, and a deviation from these constraints can predict the epistemic uncertainty of the expected outcome. Although we are focusing on end-to-end networks with implicit layers (learning to optimize strategy), one should be able to build a differentiable constraint \cite{gradslam} for end-to-end networks performing SLAM without implicit layers. In a SLAM setting, these constraints generally include minimizing geometric reprojection errors in a bundle adjustment problem \cite{tang_ba_net}, photometric errors to maintain photometric consistency \cite{yang2020d3vo}, depth map consistency errors \cite{sigma_fusion}, optical flow consistency \cite{teed_deepv2d,teed2021droid} etc. We argue that the predictive uncertainty of the network is at least the uncertainty generated from the errors imposed by constraints in implicit layers. 
\begin{equation}
   \mathbf{ \Sigma_{NN}}(\mathbf{y}) \succ \mathbf{\Sigma_{implicit}}(\mathbf{y})
\end{equation}
where $\Sigma_{NN}$ is the positive definite covariance matrix representing the uncertainty of the expected outcome $\hat{y}$ and $\Sigma_{implicit}$ is the uncertainty of the expected outcome $\hat{y}$ generated from the error $e_y$ representing the deviation from the constraint.

This approach has some distinct advantages over previous methods \cite{gal2016dropout, kendall2017uncertainties}: (1) there is a significant reduction in the number of parameters to estimate the uncertainty over drop-out monte-carlo sampling of a deep learning network with millions of parameters, and (2) we do not need to add a dropout layer in an existing network and retrain the network. Although having an implicit layer limits the generalizability of the network, the network will provide a relatively higher uncertainty estimate when given out-of-distribution samples. For example, suppose the constraint in the visual odometry network's implicit layer is designed for rigid body motion and non-deformable 3D structure. In that case, presenting an out-of-distribution sample from a deformable scene will increase the error and, therefore, the prediction uncertainty, which should be the case. It should be noted that computing uncertainty (marginal covariance) by linearizing the implicit layer error function can give only locally consistent estimates. Therefore, a factor graph formulation that works in an incremental setting \cite{isam,g2o} is an excellent tool for estimating the covariance where changes in the estimates of the variables are taken into account by their re-linearization strategies. 
\subsection{Covariance recovery with a factor graph formulation of the implicit layer}
\begin{figure}[h!]
    \captionsetup{font=footnotesize, labelfont={bf, sf}}
    \centering
    \includegraphics[width=\columnwidth]{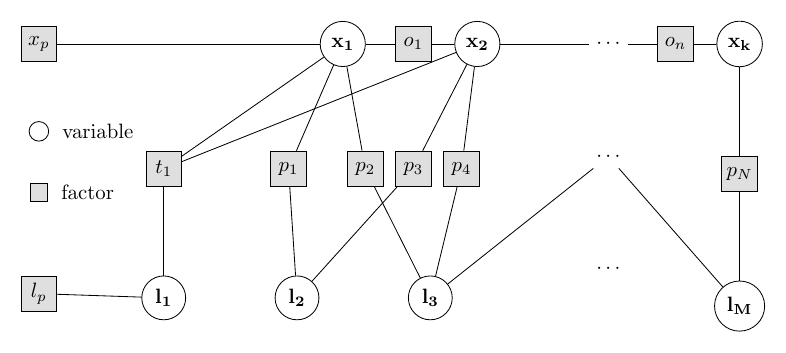}
    \caption{Factor graph representation of a bundle adjustment problem. Factors are the error residuals between connected variables represented by \eqref{eq:error_residuals}. In this diagram, $l_i$ represents landmark variables, $x_i$ represents pose variables, $t_i$ is a ternary factor with $\mathbf{X} = \{x_1, x_2, l_1\}$ , $p_i$ are binary factors with $\mathbf{X} = \{x_i, l_k\}$. like a camera projection function, and $o_i$ are binary factors with  $\mathbf{X} = \{x_i, x_k\}$, $x_p$ like an odometry function, and $l_p$ are unary factors representing prior states of variables imposing initial conditions. }
    \label{fig:factor_graph}
\end{figure}

We specifically focus on the mechanics of recovering covariance from a bundle adjustment problem. With a measurement model in place, 
\begin{equation}
\mathbf{m} = \mathbf{f(X)} + \mathbf{n}
\end{equation}
where $\mathbf{X} \subset \{\mathbf{x_1}, \mathbf{x_2}, ..., \mathbf{x_n}, \mathbf{l_1}, \mathbf{l_2}, ..., \mathbf{l_m}\} $ with Gaussian noise $\mathbf{n} \sim  \mathcal{N}(0, \mathbf{\Sigma_{i}})$. In a bundle adjustment problem, the residual functions $\mathbf{e_i}$ that depict the error between a stochastic measurement model $\mathbf{f(X)}$ and observed measurement $\mathbf{m_i^{obs}}$ are the factors in a factor graph \cite{isam} as shown in figure \cref{fig:factor_graph} 
\begin{equation}
    \mathbf{e_i} = \mathbf{f(X)} - \mathbf{m_i^{obs}} + \mathbf{n} \stackrel{linearize}\approx    \mathbf{J_i\delta X} - \mathbf{b} \label{eq:error_residuals} 
\end{equation}

The solution to the bundle adjustment problem is the iterative minimization of the non-linear least squares problem \eqref{eq:minimization} formed from the error residuals. The solution is obtained by linearizing into a least squares problem \eqref{eq:linearized} to get the most optimal delta change in state that minimizes the Mahalanobis distance cost objective. 
\begin{align}
    \mathbf{X^{*}} &= \operatorname*{argmin}_x \sum_{i=1}^{N} 
    \mathbf{e_{i}^{T}\Sigma_{i}^{-1} e_{i}}\label{eq:minimization} \\
    \stackrel{linearize}\approx \mathbf{\delta X^*} &= \operatorname*{argmin}_{\delta x} || \mathbf{A\delta X} - \mathbf{B} ||^2
    \label{eq:linearized}\\
    \mathbf{X_{new}} &= \mathbf{X_{old}} \boxplus \mathbf{\delta X^{*}}
    \label{eq:update}
\end{align}

\begin{figure}[h!]
    \captionsetup{font=footnotesize, labelfont={bf, sf}}
    \includegraphics[width=\columnwidth]{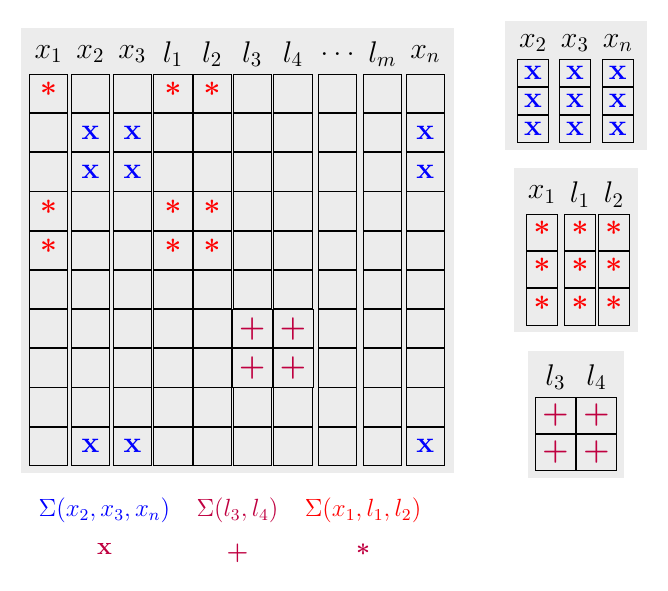}
    \caption{Recovering marginal covariances of variables is selecting respective blocks in the full covariance matrix of the bundle adjustment problem.}
    \label{fig:enter-label}
\end{figure}
where $\mathbf{A} = \Sigma^{-1/2}\mathbf{J}$, with $\mathbf{J} \in \mathbb{R}^{N\times M}$ the augmented measurement jacobian and $\Sigma$ the augmented covariance of the bundle adjustment problem at the current linearization point $X_{old}$, the augmented state vector, which can comprise poses and landmarks.
The state is then updated by retraction on the state manifold in \eqref{eq:update}. The uncertainty of the estimated states in a non-linear least squares problem is given by the delta method and is equivalent to the sparse information matrix $ \mathcal{I}(X) =  \mathbf{A^{T}A} = \mathbf{R}^T\mathbf{R} $. The inverse of the $\mathcal{I}$ is the covariance of the estimated solution 
\begin{align}
\mathbf{\Sigma} = (\mathbf{R^{T}R})^{-1}
\end{align}
It is not straightforward to recover the covariance from the information matrix. In general, recovering covariance by matrix inversion is not done, as the covariance matrix is densely populated with $n^2$ entries, which utilizes a lot of memory and results in very large computation times. But, in practice, the marginal covariance of a subset of variables is recovered from the information matrix using the Schur complement on the partitioned matrix.
\begin{align}
A &= \begin{bmatrix} A_{11} & A_{12} \\ A_{21} & A_{22} \end{bmatrix}
\text{partitioned in}  \begin{bmatrix} x_1 \\ x_2 \end{bmatrix}\\
S &= A_{11} - A_{12}A_{22}^{-1}A_{21} \implies \Sigma(x_1) = S^{-1}
\end{align}
We have used the efficient computation of marginal covariance using a recursive dynamic programming solution \cite{golub1980large, kaess2009covariance} to obtain the exact entries of the covariance matrix.

\subsection{Choice of coordinates and gauge prior on recovering consistent covariance }
We have represented poses as an element on $SE(3)$ manifold and use its associate lie algebra $\mathfrak{se}(3)$ to represent pose uncertainties. This choice of representation is free of singularities, unlike alternatives such as 3D Euler angles. In addition, this provides a minimal representation of the pose variable \cite{barfoot}. Another major advantage of representing uncertainty in the lie group space is preserving the consistency and monotonicity of uncertainty during exploration. A good choice of coordinate representation is vital for consistent covariance recovery of the poses \cite{kim_uncertainty} as no extra information is added along any direction. The optimality criteria such as D-opt, E-opt, etc. \cite{kim_uncertainty} behave well when we propagate covariances over time as new incoming frames arrive.

\begin{figure}[h!]
    \centering
    \captionsetup{font=footnotesize, labelfont={bf, sf}}
    \includegraphics[width=\columnwidth]{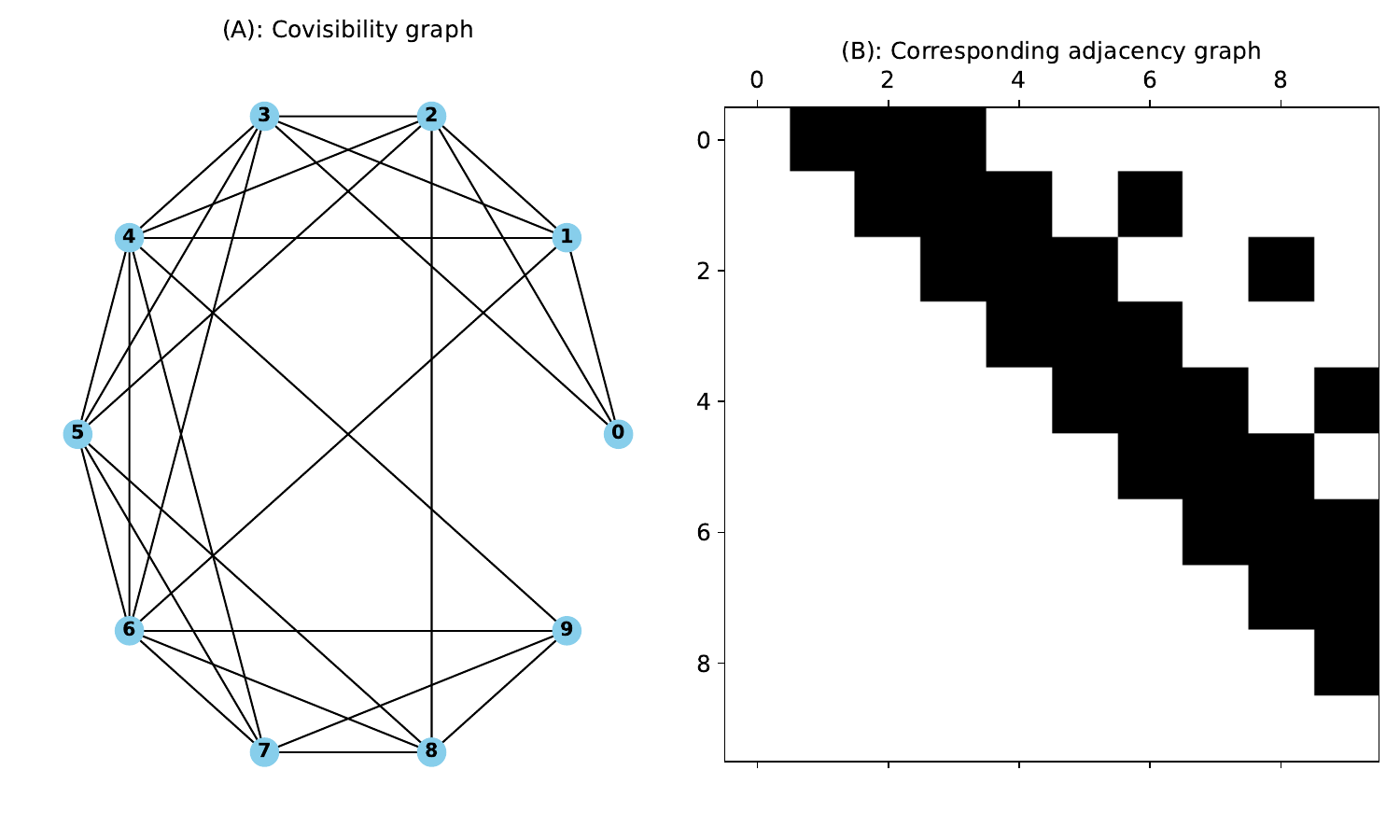}
    \caption{Covisibility graph of overlapping camera keyframes in a VO framework. The co-visibility graph, left, describes which frames have been successfully registered. The right figure depicts the co-visibility graph's adjacent matrix, showing how the current keyframe row $i$ is connected to other keyframe images $j$ at different columns. Generally, when the robot is exploring, the adjacency matrix is filled along the diagonal, making only a few connections with previous keyframes.  The co-visibility graph directly affects the set of measurements used to form the information matrix $\mathbf{A}$. So, the co-visibility graph intuitively describes how covariance should trend for any VO odometry engine.}
    \label{fig:covisibility}
\end{figure}

In an incremental Visual SLAM setting, the uncertainty is not always monotonically increasing. Generally, when the camera moves in a single direction, the incoming frames do not overlap with a few of the last keyframes. However, when the camera revisits some previously visited locations, the incoming frames can overlap with many earlier keyframes as represented by the co-visibility graph shown in \cref{fig:covisibility}. 

We have affixed the gauge freedom by setting a prior on the first two poses obtained from the monocular SLAM solution, referred to as a gauge-prior \cite{zhang_gauge}. Specifically, we add a penalty on the first two camera poses predicted from the monocular VO solution by adding the prior pose and its distribution. By doing this, we have also fixed the scale for the estimated solution, and therefore we will recover a scaled covariance in a monocular setting. 
\begin{align}
    \mathbf{X^{*}} &=  \operatorname*{argmin}_x \{||r_0||_{\Sigma_0} + ||r_1||_{\Sigma_1}  +  \sum_{i=1}^{N} 
    \mathbf{e_{i}^{T} \Sigma_{i}^{-1} e_{i}}\}\\
    r_i &= \mathbf{X_i} \boxminus \mathbf{X_i^{pred}} = (\delta \phi, \delta p) \\
\end{align}
where $X_0^{pred}$ and $X_1^{pred}$ are monocular camera predictions from the visual odometry algorithm.




\begin{figure*}[h!]
    \centering
    \includegraphics[width=0.9\textwidth]{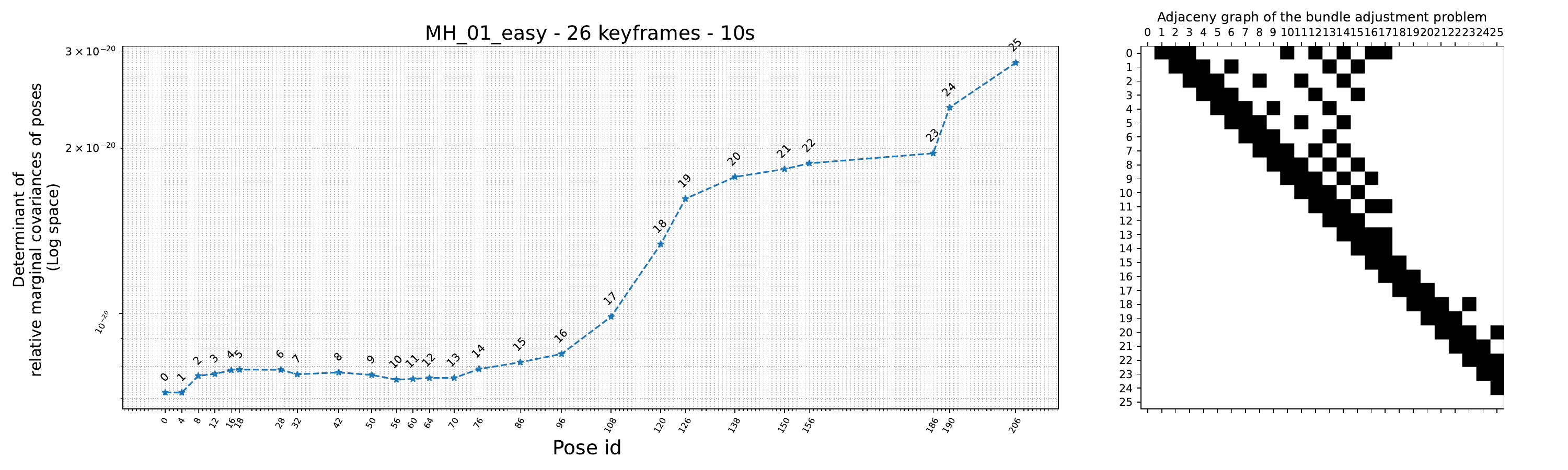}
    \label{fig:figure1}
\end{figure*}
\begin{figure*}[h!]
    \centering
    \includegraphics[width=0.9\textwidth]{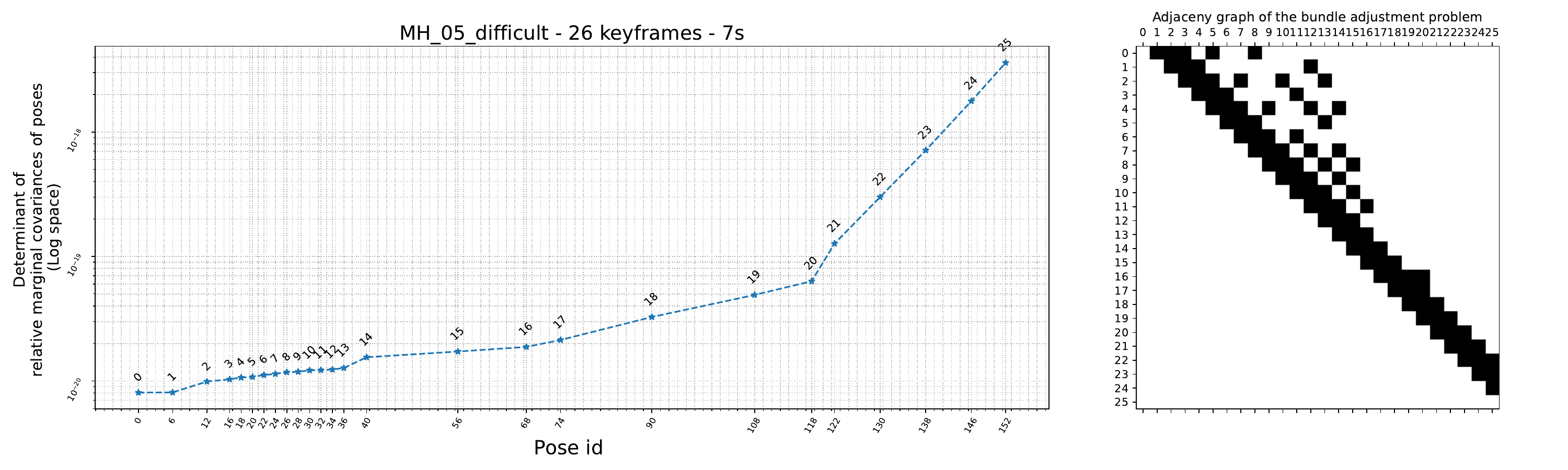}
    \label{fig:figure2}
\end{figure*}
\begin{figure*}[h!]
    \centering
    \includegraphics[width=0.9\textwidth]{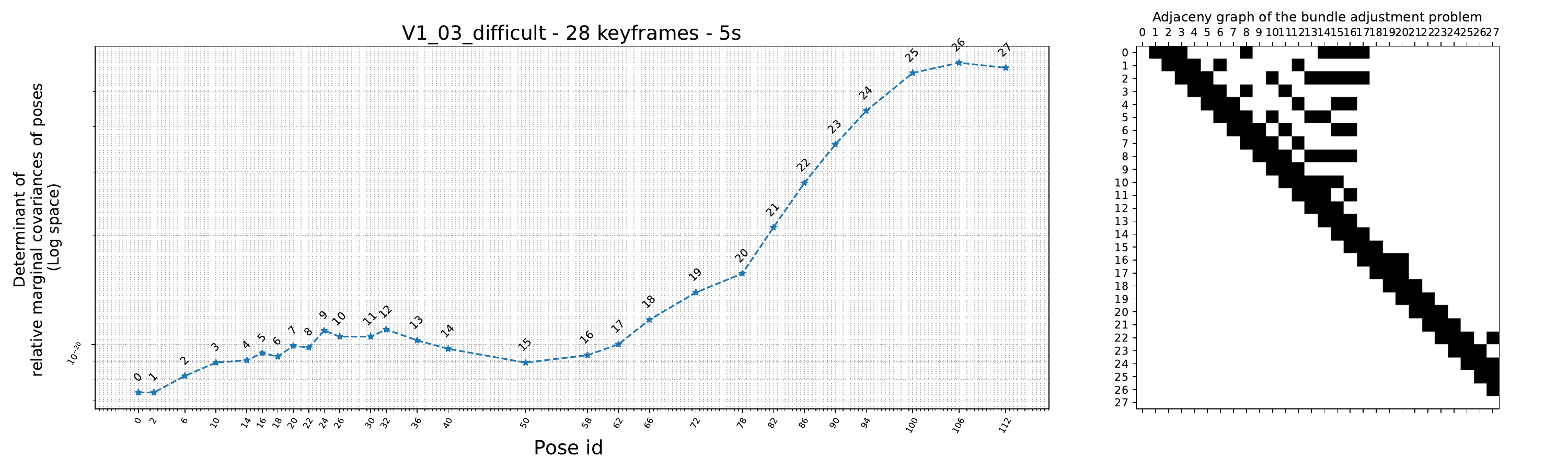}
    \label{fig:figure2}
\end{figure*}
\begin{figure*}[h!]
    \centering
    \captionsetup{font=footnotesize, labelfont={bf, sf}}
    \includegraphics[width=.9\textwidth]{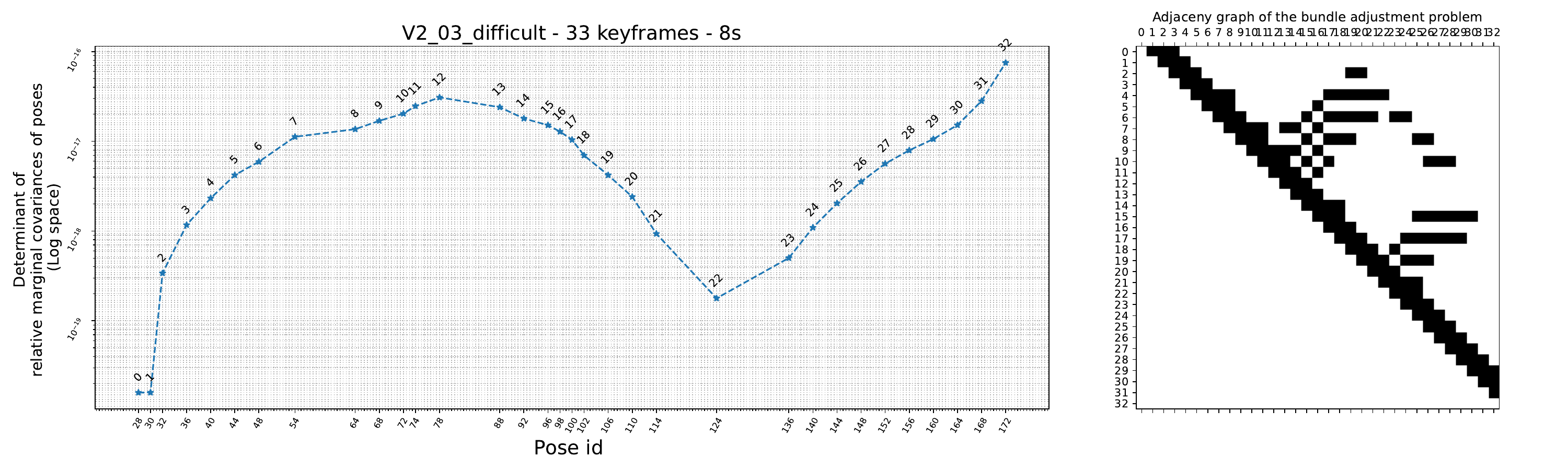}
    \caption{The left plot show the trends in marginal covariances obtained from a monocular camera VO solution obtained from a deep learning visual odometry (DROID SLAM) on some sections of EUROC \cite{euroc} datasets. The right plot shows the adjacency graph highlighting the nearest neighbors with which a keyframe overlaps and registers. Note that the marginal covariances of poses increase as the camera moves in one direction with some overlap to its previous frames. However, when the incoming frame overlaps past keyframes, indicated by off-diagonal elements on the adjacency graph, the covariance decreases as described in section \cref{sec:validation}. Also note that the rate and direction of growth of covariance directly depend on the amount of overlap. This shows that the covariance model developed using the formulation is consistent with the observed data. We have evaluated the complete datasets with similar observations throughout.}

    \label{fig:plots_covariance}
\end{figure*}
\section{Experimental Results}
In this paper, we have shown our formulation of covariance recovery on an end-to-end pre-trained deep learning visual odometry engine, DROID SLAM, \cite{teed2021droid} which has a Dense Bundle Adjustment (DBA) layer as its implicit layer. 
\begin{figure}[h!]
    \captionsetup{font=footnotesize, labelfont={bf, sf}}
    \centering
    \includegraphics[width=0.5\columnwidth]{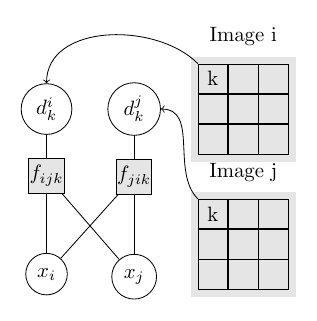}
    \caption{Ternary factor $f_{ijk}$ and $f_{jik}$ for a pair of images in the co-visibility graph in DROID SLAM. The ternary factor is an error residual of the predicted optical flow and the induced optical flow that ties the inverse depth at pixel k $d_k$, pose of camera i ($x_i$) and pose of camera j ($x_j$). The error $e_{ijk}$ in \eqref{eq:dba_error} represents the factor $f_ijk$ in the diagram}
    \label{fig:droid_factor}
\end{figure}
The DBA layer implements the error function in \eqref{eq:dba_error}, and its equivalent factor graph implementation is shown in \cref{fig:droid_factor}. $\Pi$ and $\Pi^{-1}$ are camera projection and inverse projection functions, $T_{c_i}^{w}$ and $T_{c_j}^{w}$ are respective camera to world transformations of keyframe $i$ and $j$.
\begin{equation}
    e_{ijk} = p^{*} - \Pi_{j}(\mathbf{{T}^{w}_{c_j}}^{-1} \mathbf{T}^{w}_{c_i} \Pi_{i} ^{-1}(p_{i}, d^{i}_{k}))
    \label{eq:dba_error}
\end{equation}
Intuitively, the DBA layer imposes on the network constraints that minimize the error between predicted optical flow $p^{*}$ and projected optical flow between camera poses $i$ and $j$. In Droid SLAM, the keyframes are connected from $i$ to $j$ and vice versa. However, we only show the upper triangular part for convenience in co-visibility graphs in \cref{fig:plots_covariance}.
Droid SLAM outputs predictions of the poses $T_{c_t}^w$ and a dense depth estimate $d^k_t$ of the image at time $t$. We use this optimal estimate as our linearization point at time $t$ for marginal covariance recovery from the equivalent factor graph of the DBA layer. The complete derivation of analytical jacobians and their verification is part of our  \footnote{\url{https://github.com/jpsnir/droid_slam_covariance_models}}{github repository}.

\subsection{Uncertainty measure and evaluation for consistency}

\label{sec:validation}
We have chosen the D-opt criteria to represent the growth of uncertainty over time. This implies that the uncertainty is proportional to the determinant of the marginal covariance matrix of pose at any time. The determinant captures the volume of the 6D ellipsoid information in $\mathfrak{se}(3)$ space. In \cite{kim_uncertainty}, Kim proves that the choice of criteria does not impact the trend of the uncertainty growth when variables are parameterized in the $SE(3)$ space. With the selection of D-opt criteria, the trends in uncertainty depend only on the set of measurements and predicted estimates by the nodes in the network. They should show a correlation with visual data alone. For the model validation, we have shown the covariance D-opt plots on the EUROC \cite{euroc} dataset in \cref{fig:plots_covariance} on a fixed set of keyframes obtained from DROID SLAM. We have evaluated all the EUROC datasets, but in this paper, we show only a brief section of the evaluated dataset to showcase the main results. 
\begin{figure}[h!]
    \centering
    \captionsetup{font=footnotesize, labelfont={bf, sf}}
    \includegraphics[width=\columnwidth]{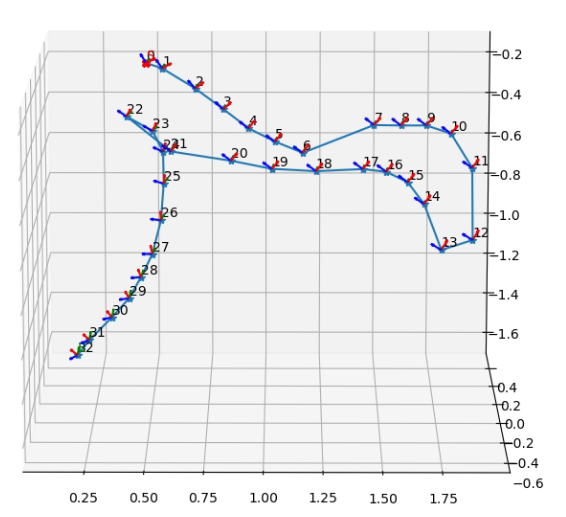}
    \caption{Front view of the v2 - 03 trajectory showing that revisiting places correlates with a decrease in covariance. Keyframes 0 to 11 have sequential overlap. After 11, from keyframe 12 to 21, the camera faces the same direction but moves in the opposite direction, producing registration with previous keyframes in the co-visibility graph window. From 22 to 23, the camera moves down and there is only sequential overlap with previous keyframes. Red(0) is the starting keyframe, and green (32) is the last keyframe }
    \label{fig:traj_v2_03}
\end{figure}

        
         
        

\subsection{Discussions}
In \cref{fig:plots_covariance}, we show the trends in marginal covariances derived from a monocular camera Visual Odometry (VO) solution, provided by DROID SLAM, applied to select sections of the EUROC datasets \cite{euroc}. On the right, the adjacency graph highlights keyframe overlaps and registrations with nearest neighbors. The marginal covariances of poses increase as the camera moves unidirectionally, with some overlap in previous frames as MH01-easy and MH05-difficult datasets. Conversely, the covariance decreases when the incoming frame aligns more closely with past keyframes, resulting in higher overlap, as depicted in the adjacency graph, as detailed in Section IV-A. This can be verified with the trajectory plot of \cref{fig:traj_v2_03}. Notably, the rate and direction of covariance growth directly correlate with the amount of overlap, indicating the consistency of the covariance model developed using our formulation with observed data. We have consistently observed similar trends throughout our evaluation of the complete EUROC datasets. 


\section{Conclusion and Future work}

The paper presents a novel method for recovering covariance from a pre-trained deep neural network by leveraging its implicit layer. The work provides empirical evidence that implicit layers can be used to model the scaled marginal covariance of poses from a monocular camera visual odometry engine. With poses parameterized in the manifold space and uncertainty in the lie algebra space, the recovered pose covariances strongly correlate with the co-visibility graph of keyframes for visual data.
Future work should include applying recovered covariances from deep-learning visual odometry engines to close large loops in a SLAM setting. Recovered covariances could also enable meaningful sensor fusion between a deep learning engine and sensors such as an IMU or LiDAR.




\printbibliography

@manual{skydio,
  title  = "skydio-drones",
  author = "",
  note   = "\url{https://www.skydio.com/}",
  year   = "accessed Feb, 2024"
}

@manual{waymo,
  title  = "waymo-av",
  author = "",
  note   = "\url{https://www.waymo.com/}",
  year   = "accessed Feb, 2024"
}

@article{sunderhauf2018limits,
  title={The limits and potentials of deep learning for robotics},
  author={S{\"u}nderhauf, Niko and Brock, Oliver and Scheirer, Walter and Hadsell, Raia and Fox, Dieter and Leitner, J{\"u}rgen and Upcroft, Ben and Abbeel, Pieter and Burgard, Wolfram and Milford, Michael and others},
  journal={The International journal of robotics research},
  volume={37},
  number={4-5},
  pages={405--420},
  year={2018},
  publisher={SAGE Publications Sage UK: London, England}
}

@article{zhan-2021-uncer-quant,
  author =       {Ni Zhan and John R. Kitchin},
  title =        {Uncertainty Quantification in Machine Learning and Nonlinear
                  Least Squares Regression Models},
  journal =      {AIChE Journal},
  volume =       {},
  number =       {},
  pages =        {},
  year =         2021,
  DATE_ADDED =   {Mon Nov 8 08:51:21 2021},
}

@article{gal2016uncertainty,
  title={Uncertainty in deep learning},
  author={Gal, Yarin and others},
  year={2016},
  publisher={phd thesis, University of Cambridge}
}

@inproceedings{gal2016dropout,
  title={Dropout as a bayesian approximation: Representing model uncertainty in deep learning},
  author={Gal, Yarin and Ghahramani, Zoubin},
  booktitle={international conference on machine learning},
  pages={1050--1059},
  year={2016},
  organization={PMLR}
}

@article{euroc,
author = {Burri, Michael and Nikolic, Janosch and Gohl, Pascal and Schneider, Thomas and Rehder, Joern and Omari, Sammy and Achtelik, Markus W and Siegwart, Roland}, 
title = {The EuRoC micro aerial vehicle datasets},
year = {2016}, 
journal = {The International Journal of Robotics Research} 
}

@article{russell2021multivariate,
  title={Multivariate uncertainty in deep learning},
  author={Russell, Rebecca L and Reale, Christopher},
  journal={IEEE Transactions on Neural Networks and Learning Systems},
  volume={33},
  number={12},
  pages={7937--7943},
  year={2021},
  publisher={IEEE}
}

@article{hullermeier2021aleatoric,
  title={Aleatoric and epistemic uncertainty in machine learning: An introduction to concepts and methods},
  author={H{\"u}llermeier, Eyke and Waegeman, Willem},
  journal={Machine Learning},
  volume={110},
  pages={457--506},
  year={2021},
  publisher={Springer}
}

@article{loquercio2020general,
  title={A general framework for uncertainty estimation in deep learning},
  author={Loquercio, Antonio and Segu, Mattia and Scaramuzza, Davide},
  journal={IEEE Robotics and Automation Letters},
  volume={5},
  number={2},
  pages={3153--3160},
  year={2020},
  publisher={IEEE}
}

@article{Czarnowski_deep_factors,
   author = {Czarnowski, J and Laidlow, T and Clark, R and Davison, AJ},
   journal = {IEEE Robotics and Automation Letters},
   pages = {721--728},
   title = {DeepFactors: Real-time probabilistic dense monocular SLAM},
   volume = {5},
   year = {2020}
}

@article{kendall2017uncertainties,
  title={What uncertainties do we need in bayesian deep learning for computer vision?},
  author={Kendall, Alex and Gal, Yarin},
  journal={Advances in neural information processing systems},
  volume={30},
  year={2017}
}

@INPROCEEDINGS{katherince_covest,
  author={Liu, Katherine and Ok, Kyel and Vega-Brown, William and Roy, Nicholas},
  booktitle={2018 IEEE International Conference on Robotics and Automation (ICRA)}, 
  title={Deep Inference for Covariance Estimation: Learning Gaussian Noise Models for State Estimation}, 
  year={2018},
  volume={},
  number={},
  pages={1436-1443},
}

@article{teed_deepv2d,
  title={Deepv2d: Video to depth with differentiable structure from motion},
  author={Teed, Zachary and Deng, Jia},
  journal={arXiv preprint arXiv:1812.04605},
  year={2018}
}

@InProceedings{zhou_deeptam,
    author       = "H. Zhou and B. Ummenhofer and T. Brox",
    title        = "DeepTAM: Deep Tracking and Mapping",
    booktitle    = "European Conference on Computer Vision (ECCV)",
    month        = " ",
    year         = "2018"
}

@inproceedings{bloesch2018codeslam,
  title={Codeslam—learning a compact, optimisable representation for dense visual slam},
  author={Bloesch, Michael and Czarnowski, Jan and Clark, Ronald and Leutenegger, Stefan and Davison, Andrew J},
  booktitle={Proceedings of the IEEE conference on computer vision and pattern recognition},
  pages={2560--2568},
  year={2018}
}

@inproceedings{ummenhofer2017demon,
  title={Demon: Depth and motion network for learning monocular stereo},
  author={Ummenhofer, Benjamin and Zhou, Huizhong and Uhrig, Jonas and Mayer, Nikolaus and Ilg, Eddy and Dosovitskiy, Alexey and Brox, Thomas},
  booktitle={Proceedings of the IEEE conference on computer vision and pattern recognition},
  pages={5038--5047},
  year={2017}
}

@article{tang_ba_net,
  author       = {Chengzhou Tang and
                  Ping Tan},
  title        = {BA-Net: Dense Bundle Adjustment Network},
  journal      = {CoRR},
  volume       = {abs/1806.04807},
  year         = {2018},
  bibsource    = {dblp computer science bibliography, https://dblp.org}
}

@article{zihan_niceslam,
  title={NICE-SLAM: Neural Implicit Scalable Encoding for SLAM},
  author={Zihan Zhu and Songyou Peng and Viktor Larsson and Weiwei Xu and Hujun Bao and Zhaopeng Cui and Martin R. Oswald and Marc Pollefeys},
  journal={2022 IEEE/CVF Conference on Computer Vision and Pattern Recognition (CVPR)},
  year={2021},
  pages={12776-12786},
}

@article{tartanvo2020corl,
  title =   {TartanVO: A Generalizable Learning-based VO},
  author =  {Wang, Wenshan and Hu, Yaoyu and Scherer, Sebastian},
  booktitle = {Conference on Robot Learning (CoRL)},
  year =    {2020}
}

@inproceedings{yang2020d3vo,
  title={D3vo: Deep depth, deep pose and deep uncertainty for monocular visual odometry},
  author={Yang, Nan and Stumberg, Lukas von and Wang, Rui and Cremers, Daniel},
  booktitle={Proceedings of the IEEE/CVF conference on computer vision and pattern recognition},
  pages={1281--1292},
  year={2020}
}

@INPROCEEDINGS{nerf_slam,
  author={Rosinol, Antoni and Leonard, John J. and Carlone, Luca},
  booktitle={2023 IEEE/RSJ International Conference on Intelligent Robots and Systems (IROS)}, 
  title={NeRF-SLAM: Real-Time Dense Monocular SLAM with Neural Radiance Fields}, 
  year={2023},
  volume={},
  number={},
  pages={3437-3444},
}

@inproceedings{sigma_fusion,
  title={Probabilistic volumetric fusion for dense monocular slam},
  author={Rosinol, Antoni and Leonard, John J and Carlone, Luca},
  booktitle={Proceedings of the IEEE/CVF Winter Conference on Applications of Computer Vision},
  pages={3097--3105},
  year={2023}
}

@inproceedings{amos2017optnet,
  title={Optnet: Differentiable optimization as a layer in neural networks},
  author={Amos, Brandon and Kolter, J Zico},
  booktitle={International Conference on Machine Learning},
  pages={136--145},
  year={2017},
  organization={PMLR}
}

@phdthesis{teed_thesis,
author={Teed,Zachary},
year={2022},
title={Optimization Inspired Neural Networks for Multiview 3D Reconstruction},
journal={ProQuest Dissertations and Theses},
pages={105},
note={Copyright - Database copyright ProQuest LLC; ProQuest does not claim copyright in the individual underlying works; Last updated - 2023-03-08},
isbn={9798352694114},
language={English},

}

@article{gradslam,
    author  = { {Krishna Murthy}, Jatavallabhula and Saryazdi, Soroush and Iyer, Ganesh and Paull, Liam },
    title   = { gradSLAM: Dense SLAM meets Automatic Differentiation },
    journal = { arXiv },
    year    = { 2020 },
}

@article{teed2021droid,
  title={Droid-slam: Deep visual slam for monocular, stereo, and rgb-d cameras},
  author={Teed, Zachary and Deng, Jia},
  journal={Advances in neural information processing systems},
  volume={34},
  pages={16558--16569},
  year={2021}
}

@article{ORBSLAM3_TRO,
  title={{ORB-SLAM3}: An Accurate Open-Source Library for Visual, Visual-Inertial 
           and Multi-Map {SLAM}},
  author={Campos, Carlos AND Elvira, Richard AND G\´omez, Juan J. AND Montiel, 
          Jos\'e M. M. AND Tard\'os, Juan D.},
  journal={IEEE Transactions on Robotics}, 
  volume={37},
  number={6},
  pages={1874-1890},
  year={2021}
 }

@ARTICLE{vins_mono,
  author={Qin, Tong and Li, Peiliang and Shen, Shaojie},
  journal={IEEE Transactions on Robotics}, 
  title={VINS-Mono: A Robust and Versatile Monocular Visual-Inertial State Estimator}, 
  year={2018},
  volume={34},
  number={4},
  pages={1004-1020},
  }

@InProceedings{Kimera,
  title = {Kimera: an Open-Source Library for Real-Time Metric-Semantic Localization and Mapping},
  author = {Rosinol, Antoni and Abate, Marcus and Chang, Yun and Carlone, Luca},
  year = {2020},
  booktitle = {IEEE Intl. Conf. on Robotics and Automation (ICRA)},
  
}

@article{golub1980large,
  title={Large-scale geodetic least-squares adjustment by dissection and orthogonal decomposition},
  author={Golub, Gene H and Plemmons, Robert J},
  journal={Linear Algebra and Its Applications},
  volume={34},
  pages={3--28},
  year={1980},
  publisher={Elsevier}
}

@ARTICLE{zhang_gauge,
  author={Zhang, Zichao and Gallego, Guillermo and Scaramuzza, Davide},
  journal={IEEE Robotics and Automation Letters}, 
  title={On the Comparison of Gauge Freedom Handling in Optimization-Based Visual-Inertial State Estimation}, 
  year={2018},
  volume={3},
  number={3},
  pages={2710-2717},
 }

@INPROCEEDINGS{kim_uncertainty,
  author={Kim, Youngji and Kim, Ayoung},
  booktitle={2017 IEEE/RSJ International Conference on Intelligent Robots and Systems (IROS)}, 
  title={On the uncertainty propagation: Why uncertainty on lie groups preserves monotonicity?}, 
  year={2017},
  volume={},
  number={},
  pages={3425-3432},
  }

@INPROCEEDINGS{isam,
  author={Kaess, Michael and Johannsson, Hordur and Roberts, Richard and Ila, Viorela and Leonard, John and Dellaert, Frank},
  booktitle={2011 IEEE International Conference on Robotics and Automation}, 
  title={iSAM2: Incremental smoothing and mapping with fluid relinearization and incremental variable reordering}, 
  year={2011},
  volume={},
  number={},
  pages={3281-3288},
  }

@inproceedings{g2o,
  title={g2o: A general framework for (hyper) graph optimization},
  author={Grisetti, Giorgio and K{\"u}mmerle, Rainer and Strasdat, Hauke and Konolige, Kurt},
  booktitle={Proceedings of the IEEE International Conference on Robotics and Automation (ICRA)},
  pages={9--13},
  year={2011}
}

@book{prob_rob,
author = {Thrun, Sebastian and Burgard, Wolfram and Fox, Dieter},
title = {Probabilistic Robotics (Intelligent Robotics and Autonomous Agents)},
year = {2005},
isbn = {0262201623},
publisher = {The MIT Press}
}

@article{poddar2017comprehensive,
  title={A comprehensive overview of inertial sensor calibration techniques},
  author={Poddar, Shashi and Kumar, Vipan and Kumar, Amod},
  journal={Journal of Dynamic Systems, Measurement, and Control},
  volume={139},
  number={1},
  pages={011006},
  year={2017},
  publisher={American Society of Mechanical Engineers}
}

@inproceedings{song2013survey,
  title={Survey on camera calibration technique},
  author={Song, Liming and Wu, Wenfu and Guo, Junrong and Li, Xiuhua},
  booktitle={2013 5th International conference on intelligent human-machine systems and cybernetics},
  volume={2},
  pages={389--392},
  year={2013},
  organization={IEEE}
}

@inproceedings{wang2021survey,
  title={A survey of extrinsic calibration of lidar and camera},
  author={Wang, Yuru and Li, Jian and Sun, Yi and Shi, Meiping},
  booktitle={International Conference on Autonomous Unmanned Systems},
  pages={933--944},
  year={2021},
  organization={Springer}
}

@article{kaess2009covariance,
  title={Covariance recovery from a square root information matrix for data association},
  author={Kaess, Michael and Dellaert, Frank},
  journal={Robotics and autonomous systems},
  volume={57},
  number={12},
  pages={1198--1210},
  year={2009},
  publisher={Elsevier}
}

@ARTICLE{barfoot,
  author={Barfoot, Timothy D. and Furgale, Paul T.},
  journal={IEEE Transactions on Robotics}, 
  title={Associating Uncertainty With Three-Dimensional Poses for Use in Estimation Problems}, 
  year={2014},
  volume={30},
  number={3},
  pages={679-693},
  }
\end{document}